# An Improved Parser for Data-Oriented Lexical-Functional Analysis

**Rens Bod**

Informatics Research Institute, University of Leeds, Leeds LS2 9JT, UK, &
Institute for Logic, Language and Computation, University of Amsterdam
rens@scs.leeds.ac.uk

**Abstract**
We present an LFG-DOP parser which uses fragments from LFG-annotated sentences to parse new sentences. Experiments with the Verbmobil and Homecentre corpora show that (1) Viterbi *n* best search performs about 100 times faster than Monte Carlo search while both achieve the same accuracy; (2) the DOP hypothesis which states that parse accuracy increases with increasing fragment size is confirmed for LFG-DOP; (3) LFG-DOP's relative frequency estimator performs worse than a discounted frequency estimator; and (4) LFG-DOP significantly outperforms Tree-DOP if evaluated on tree structures only.

## 1 Introduction

Data-Oriented Parsing (DOP) models learn how to provide linguistic representations for an unlimited set of utterances by generalizing from a given corpus of properly annotated exemplars. They operate by decomposing the given representations into (arbitrarily large) fragments and recomposing those pieces to analyze new utterances. A probability model is used to choose from the collection of different fragments of different sizes those that make up the most appropriate analysis of an utterance.

DOP models have been shown to achieve state-of-the-art parsing performance on benchmarks such as the Wall Street Journal corpus (see Bod 2000a). The original DOP model in Bod (1993) was based on utterance analyses represented as surface trees, and is equivalent to a Stochastic Tree-Substitution Grammar. But the model has also been applied to several other grammatical frameworks, e.g. Tree-Insertion Grammar (Hoogweg 2000), Tree-Adjoining Grammar (Neumann 1998), Lexical-Functional Grammar (Bod & Kaplan 1998; Cormons 1999), Head-driven Phrase Structure Grammar (Neumann & Flickinger 1999), and Montague Grammar (Bonnema et al. 1997; Bod 1999). Most probability models for DOP use the relative frequency estimator to estimate fragment probabilities, although Bod (2000b) trains fragment probabilities by a maximum likelihood reestimation procedure belonging to the class of expectation-maximization algorithms. The DOP model has also been tested as a model for human sentence processing (Bod 2000d).

This paper presents ongoing work on DOP models for Lexical-Functional Grammar representations, known as LFG-DOP (Bod & Kaplan 1998). We develop a parser which uses fragments from LFG-annotated sentences to parse new sentences, and we derive some experimental properties of LFG-DOP on two LFG-annotated corpora: the Verbmobil and Homecentre corpus. The experiments show that the DOP hypothesis, which states that there is an increase in parse accuracy if larger fragments are taken into account (Bod 1998), is confirmed for LFG-DOP. We report on an improved search technique for estimating the most probable analysis. While a Monte Carlo search converges provably to the most probable parse, a Viterbi *n* best search performs as well as Monte Carlo while its processing time is two orders of magnitude faster. We also show that LFG-DOP outperforms Tree-DOP if evaluated on tree structures only.

## 2 Summary of LFG-DOP

In accordance with Bod (1998), a particular DOP model is described by

- a definition of a well-formed *representation for utterance analyses,*
- a set of *decomposition operations* that divide a given utterance analysis into a set of *fragments,*

- a set of *composition operations* by which such fragments may be recombined to derive an analysis of a new utterance, and
- a definition of a *probability model* that indicates how the probability of a new utterance analysis is computed.

In defining a DOP model for LFG representations, Bod & Kaplan (1998) give the following settings for DOP's four parameters.

## 2.1 Representations

The representations used by LFG-DOP are directly taken from LFG: they consist of a c-structure, an f-structure and a mapping φ between them. The following figure shows an example representation for *Kim eats*. (We leave out some features to keep the example simple.)

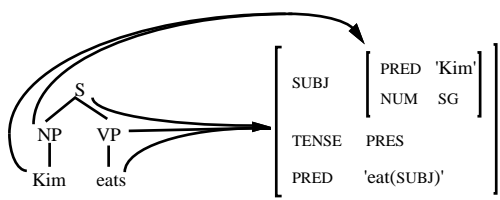

Figure 1

Bod & Kaplan also introduce the notion of accessibility which they later use for defining the decomposition operations of LFG-DOP:

An f-structure unit *f* is φ-*accessible* from a node *n* iff either *n* is φ-linked to *f* (that is, *f* = φ(*n*) ) or *f* is contained within φ(*n*) (that is, there is a chain of attributes that leads from φ(*n*) to *f*).

According to the LFG representation theory, c-structures and f-structures must satisfy certain formal well-formedness conditions. A c-structure/f-structure pair is a *valid* LFG representation only if it satisfies the Nonbranching Dominance, Uniqueness, Coherence and Completeness conditions (Kaplan & Bresnan 1982).

## 2.2 Decomposition operations and Fragments

The fragments for LFG-DOP consist of connected subtrees whose nodes are in φ-correspondence with the correponding sub-units of f-structures. To give a precise definition of LFG-DOP fragments, it is convenient to recall the decomposition operations employed by the orginal DOP model which is also known as the "Tree-DOP" model (Bod 1993, 1998):

(1) *Root*: the *Root* operation selects any node of a tree to be the root of the new subtree and erases all nodes except the selected node and the nodes it dominates.

(2) *Frontier*: the *Frontier* operation then chooses a set (possibly empty) of nodes in the new subtree different from its root and erases all subtrees dominated by the chosen nodes.

Bod & Kaplan extend Tree-DOP's *Root* and *Frontier* operations so that they also apply to the nodes of the c-structure in LFG, while respecting the principles of c/f-structure correspondence.

When a node is selected by the *Root* operation, all nodes outside of that node's subtree are erased, just as in Tree-DOP. Further, for LFG-DOP, all φ links leaving the erased nodes are removed and all f-structure units that are not φ-accessible from the remaining nodes are erased. For example, if *Root* selects the NP in figure 1, then the f-structure corresponding to the S node is erased, giving figure 2 as a possible fragment:

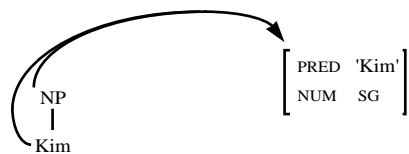

Figure 2

In addition the *Root* operation deletes from the remaining f-structure all semantic forms that are local to f-structures that correspond to erased c-structure nodes, and it thereby also maintains the fundamental two-way connection between words and meanings. Thus, if *Root* selects the VP node so that the NP is erased, the subject semantic form "Kim" is also deleted:

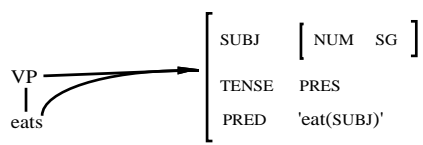

Figure 3

As with Tree-DOP, the *Frontier* operation then selects a set of frontier nodes and deletes all subtrees they dominate. Like *Root*, it also removes the φ links of the deleted nodes and erases any semantic form that corresponds to any of those nodes. For instance, if the NP in figure 1 is selected as a frontier node, *Frontier* erases the predicate "Kim" from the fragment:

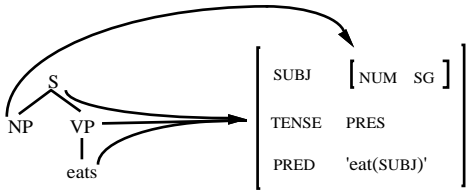

Figure 4

Finally, Bod & Kaplan present a third decomposition operation, *Discard*, defined to construct generalizations of the fragments supplied by *Root* and *Frontier*. *Discard* acts to delete combinations of attribute-value pairs subject to the following condition: *Discard* does not delete pairs whose values φ-correspond to remaining c-structure nodes. According to Bod & Kaplan (1998), *Discard*-generated fragments are needed to parse sentences that are "ungrammatical with respect to the corpus", thus increasing the robustness of the model.

### 2.3 The composition operation

In LFG-DOP the operation for combining fragments is carried out in two steps. First the c-structures are combined by leftmost substitution subject to the category-matching condition, as in Tree-DOP. This is followed by the recursive unification of the f-structures corresponding to the matching nodes. A derivation for an LFG-DOP representation $R$ is a sequence of fragments the first of which is labeled with $S$ and for which the iterative application of the composition operation produces $R$. For an illustration of the composition operation, see Bod & Kaplan (1998).

### 2.4 Probability models

As in Tree-DOP, an LFG-DOP representation $R$ can typically be derived in many different ways. If each derivation $D$ has a probability $P(D)$, then the probability of deriving $R$ is the sum of the individual derivation probabilities:

(1)    $P(R) = \Sigma_{D \text{ derives } R} P(D)$

An LFG-DOP derivation is produced by a stochastic process which starts by randomly choosing a fragment whose c-structure is labeled with the initial category. At each subsequent step, a next fragment is chosen at random from among the fragments that can be composed with the current subanalysis. The chosen fragment is composed with the current subanalysis to produce a new one; the process stops when an analysis results with no non-terminal leaves. We will call the set of composable fragments at a certain step in the stochastic process the *competition set* at that step. Let $CP(f \mid CS)$ denote the probability of choosing a fragment $f$ from a competition set CS containing $f$, then the probability of a derivation $D = <f_1, f_2 ... f_k>$ is

(2)    $P(<f_1, f_2 ... f_k>) = \Pi_i CP(f_i \mid CS_i)$

where the *competition probability* $CP(f \mid CS)$ is expressed in terms of fragment probabilities $P(f)$:

(3)    $CP(f \mid CS) = \dfrac{P(f)}{\Sigma_{f' \in CS} P(f')}$

Bod & Kaplan give three definitions of increasing complexity for the competition set: the first definition groups all fragments that only satisfy the Category-matching condition of the composition operation; the second definition groups all fragments which satisfy both Category-matching and Uniqueness; and the third definition groups all fragments which satisfy Category-matching, Uniqueness and Coherence. Bod & Kaplan point out that the Completeness condition cannot be enforced at each step of the stochastic derivation process, and is a property of the final representation which can only be enforced by sampling valid representations from the output of the stochastic process. In this paper, we will only deal with the third definition of competition set, as it selects only those fragments at each derivation step that may finally result into a valid LFG representation, thus reducing the off-line validity checking to the Completeness condition.

Note that the computation of the competition probability in the above formulas still requires a definition for the fragment probability P(*f*). Bod and Kaplan define the probability of a fragment simply as its relative frequency in the bag of all fragments generated from the corpus, just as in most Tree-DOP models. We will refer to this fragment estimator as "simple relative frequency" or "simple RF".

We will also use an alternative definition of fragment probability which is a refinement of simple RF. This alternative fragment probability definition distinguishes between fragments supplied by *Root*/*Frontier* and fragments supplied by *Discard*. We will treat the first type of fragments as seen events, and the second type of fragments as previously unseen events. We thus create two separate bags corresponding to two separate distributions: a bag with fragments generated by *Root* and *Frontier*, and a bag with fragments generated by *Discard*. We assign probability mass to the fragments of each bag by means of *discounting*: the relative frequencies of seen events are discounted and the gained probability mass is reserved for the bag of unseen events (cf. Ney et al. 1997). We accomplish this by a very simple estimator: the Turing-Good estimator (Good 1953) which computes the probability mass of unseen events as $n_1/N$ where $n_1$ is the number of singleton events and $N$ is the total number of seen events. This probability mass is assigned to the bag of *Discard*-generated fragments. The remaining mass $(1 - n_1/N)$ is assigned to the bag of *Root*/*Frontier*-generated fragments. The probability of each fragment is then computed as its relative frequency in its bag multiplied by the probability mass assigned to this bag. Let $|f|$ denote the frequency of a fragment *f*, then its probability is given by:

(4) $\quad$ P(*f* | *f* is generated by *Root*/*Frontier*) =

$$(1 - n_1/N) \frac{|f|}{\sum_{f': f' \text{ is generated by } Root/Frontier} |f'|}$$

(5) $\quad$ P(*f* | *f* is generated by *Discard*) =

$$(n_1/N) \frac{|f|}{\sum_{f': f' \text{ is generated by } Discard} |f'|}$$

We will refer to this fragment probability estimator as "discounted relative frequency" or "discounted RF".

## 4 Parsing with LFG-DOP

In his PhD-thesis, Cormons (1999) presents a parsing algorithm for LFG-DOP which is based on the Tree-DOP parsing technique described in Bod (1998). Cormons first converts LFG-representations into more compact indexed trees: each node in the c-structure is assigned an index which refers to the φ-corresponding f-structure unit. For example, the representation in figure 1 is indexed as

(S.1 $\quad$ (NP.2 $\quad$ Kim.2)
$\quad\quad$ (VP.1 $\quad$ eats.1))

where

1 --> [ (SUBJ = 2)
$\quad\quad$ (TENSE = PRES)
$\quad\quad$ (PRED = eat(SUBJ)) ]

2 --> [ (PRED = Kim)
$\quad\quad$ (NUM = SG) ]

The indexed trees are then fragmented by applying the Tree-DOP decomposition operations described in section 2. Next, the LFG-DOP decomposition operations *Root*, *Frontier* and *Discard* are applied to the f-structure units that correspond to the indices in the c-structure subtrees. Having obtained the set of LFG-DOP fragments in this way, each test sentence is parsed by a bottom-up chart parser using initially the indexed subtrees only.

Thus only the Category-matching condition is enforced during the chart-parsing process. The Uniqueness and Coherence conditions of the corresponding f-structure units are enforced during the disambiguation or chart-decoding process. Disambiguation is accomplished by computing a large number of random derivations from the chart and by selecting the analysis which results most often from these derivations. This technique is known as "Monte Carlo disambiguation" and has been extensively described in the literature (e.g. Bod 1993, 1998; Chappelier & Rajman 2000; Goodman 1998; Hoogweg 2000). Sampling a random derivation from the chart consists of choosing at random one of the fragments from the set of *composable* fragments at every labeled chart-entry (where the random choices at each chart-entry are based on the probabilities of the fragments). The derivations are sampled in a top-down, leftmost order so as to maintain the LFG-DOP derivation order. Thus the competition sets of composable fragments are computed on the fly during the Monte Carlo sampling process by grouping the f-structure units that unify and that are coherent with the subderivation built so far.

As mentioned in section 3, the Completeness condition can only be checked after the derivation process. Incomplete derivations are simply removed from the sampling distribution. After sampling a sufficiently large number of random derivations that satisfy the LFG validity requirements, the most probable analysis is estimated by the analysis which results most often from the sampled derivations. As a stop condition on the number of sampled derivations, we compute the probability of error, which is the probability that the analysis that is most frequently generated by the sampled derivations is not equal to the most probable analysis, and which is set to 0.05 (see Bod 1998). In order to rule out the possibility that the sampling process never stops, we use a maximum sample size of 10,000 derivations.

While the Monte Carlo disambiguation technique converges provably to the most probable analysis, it is quite inefficient. It is possible to use an alternative, heuristic search based on Viterbi $n$ best (we will not go into the PCFG-reduction technique presented in Goodman (1998) since that heuristic only works for Tree-DOP and is beneficial only if all subtrees are taken into account and if the so-called "labeled recall parse" is computed). A Viterbi $n$ best search for LFG-DOP estimates the most probable analysis by computing $n$ most probable derivations, and by then summing up the probabilities of the valid derivations that produce the same analysis. The algorithm for computing $n$ most probable derivations follows straightforwardly from the algorithm which computes the most probable derivation by means of Viterbi optimization (see e.g. Sima'an 1999).

## 5 Experimental Evaluation

We derived some experimental properties of LFG-DOP by studying its behavior on the two LFG-annotated corpora that are currently available: the Verbmobil corpus and the Homecentre corpus. Both corpora were annotated at Xerox PARC. They contain packed LFG-representations (Maxwell & Kaplan 1991) of the grammatical parses of each sentence together with an indication which of these parses is the correct one. For our experiments we only used the correct parses of each sentence resulting in 540 Verbmobil parses and 980 Homecentre parses. Each corpus was divided into a 90% training set and a 10% test set. This division was random except for one constraint: that all the words in the test set actually occurred in the training set. The sentences from the test set were parsed and disambiguated by means of the fragments from the training set. Due to memory limitations, we restricted the maximum depth of the indexed subtrees to 4. Because of the small size of the corpora we averaged our results on 10 different training/test set splits. Besides an *exact match* accuracy metric, we also used a more fine-grained score based on the well-known PARSEVAL metrics that evaluate phrase-structure trees (Black et al. 1991). The PARSEVAL metrics compare a proposed parse $P$ with the corresponding correct treebank parse $T$ as follows:

$$\text{Precision} = \frac{\text{\# correct constituents in } P}{\text{\# constituents in } P}$$

$$\text{Recall} = \frac{\text{\# correct constituents in } P}{\text{\# constituents in } T}$$

A constituent in *P* is correct if there exists a constituent in *T* of the same label that spans the same words and that φ-corresponds to the same f-structure unit (see Bod 2000c for some illustrations of these metrics for LFG-DOP).

**5.1 Comparing the two fragment estimators**

We were first interested in comparing the performance of the simple RF estimator against the discounted RF estimator. Furthermore, we want to study the contribution of generalized fragments to the parse accuracy. We therefore created for each training set two sets of fragments: one which contains *all* fragments (up to depth 4) and one which excludes the generalized fragments as generated by *Discard*. The exclusion of these *Discard*-generated fragments means that all probability mass goes to the fragments generated by *Root* and *Frontier* in which case the two estimators are equivalent. The following two tables present the results of our experiments where +Discard refers to the full set of fragments and −Discard refers to the fragment set without *Discard*-generated fragments.

| Estimator | Exact Match | | Precision | | Recall | |
|---|---|---|---|---|---|---|
| | +Discard | −Discard | +Discard | −Discard | +Discard | −Discard |
| Simple RF | 1.1% | 35.2% | 13.8% | 76.0% | 11.5% | 74.9% |
| Discounted RF | 35.9% | 35.2% | 77.5% | 76.0% | 76.4% | 74.9% |

Table 1. Experimental results on the Verbmobil

| Estimator | Exact Match | | Precision | | Recall | |
|---|---|---|---|---|---|---|
| | +Discard | −Discard | +Discard | −Discard | +Discard | −Discard |
| Simple RF | 2.7% | 37.9% | 17.1% | 77.8% | 15.5% | 77.2% |
| Discounted RF | 38.4% | 37.9% | 80.0% | 77.8% | 78.6% | 77.2% |

Table 2. Experimental results on the Homecentre

The tables show that the simple RF estimator scores extremely bad if all fragments are used: the exact match is only 1.1% on the Verbmobil corpus and 2.7% on the Homecentre corpus, whereas the discounted RF estimator scores respectively 35.9% and 38.4% on these corpora.

Also the more fine-grained precision and recall scores obtained with the simple RF estimator are quite low: e.g. 13.8% and 11.5% on the Verbmobil corpus, where the discounted RF estimator obtains 77.5% and 76.4%. Interestingly, the accuracy of the simple RF estimator is much higher if *Discard*-generated fragments are excluded. This suggests that treating generalized fragments probabilistically in the same way as ungeneralized fragments is harmful.

The tables also show that the inclusion of *Discard*-generated fragments leads only to a slight accuracy increase under the discounted RF estimator. Unfortunately, according to paired *t*-testing only the differences for the precision scores on the Homecentre corpus were statistically significant.

**5.2 Comparing different fragment sizes**

We were also interested in the impact of fragment size on the parse accuracy. We therefore performed a series of experiments where the fragment set is restricted to fragments of a certain maximum depth (where the depth of a fragment is defined as the longest path from root to leaf of its c-structure unit). We used the same training/test set splits as in the previous experiments and used both ungeneralized and generalized fragments together with the discounted RF estimator.

| Fragment Depth | Exact Match | Precision | Recall |
|---|---|---|---|
| 1 | 30.6% | 74.2% | 72.2% |
| ≤2 | 34.1% | 76.2% | 74.5% |
| ≤3 | 35.6% | 76.8% | 75.9% |
| ≤4 | 35.9% | 77.5% | 76.4% |

Table 3. Accuracies on the Verbmobil

| Fragment Depth | Exact Match | Precision | Recall |
|---|---|---|---|
| 1 | 31.3% | 75.0% | 71.5% |
| ≤2 | 36.3% | 77.1% | 74.7% |
| ≤3 | 37.8% | 77.8% | 76.1% |
| ≤4 | 38.4% | 80.0% | 78.6% |

Table 4. Accuracies on the Homecentre

Tables 3 and 4 show that there is a consistent increase in parse accuracy for all metrics if larger fragments are included, but that the increase itself decreases. This phenomenon is also known as the DOP hypothesis (Bod 1998), and has been confirmed for Tree-DOP on the ATIS, OVIS and Wall Street Journal treebanks (see Bod 1993, 1998, 1999, 2000a; Sima'an 1999; Bonnema et al. 1997; Hoogweg 2000). The current result thus extends the validity of the DOP hypothesis to LFG annotations. We do not yet know whether the accuracy continues to increase if even larger fragments are included (for Tree-DOP it has been shown that the accuracy *decreases* after a certain depth, probably due to overfitting -- cf. Bonnema et al. 1997; Bod 2000a).

### 5.3 Comparing LFG-DOP to Tree-DOP

In the following experiment, we are interested in the impact of functional structures on predicting the correct tree structures. We therefore removed all f-structure units from the fragments, thus yielding a Tree-DOP model, and compared the results against the full LFG-DOP model (using the discounted RF estimator and all fragments up to depth 4). We evaluated the parse accuracy on the tree structures only, using exact match together with the standard PARSEVAL measures. We used the same training/test set splits as in the previous experiments.

| Model | Exact Match | Precision | Recall |
|---|---|---|---|
| Tree-DOP | 46.6% | 88.9% | 86.7% |
| LFG-DOP | 50.8% | 90.3% | 88.4% |

Table 5. Tree accuracy on the Verbmobil

| Model | Exact Match | Precision | Recall |
|---|---|---|---|
| Tree-DOP | 49.0% | 93.4% | 92.1% |
| LFG-DOP | 53.2% | 95.8% | 94.7% |

Table 6. Tree accuracy on the Homecentre

The results indicate that LFG-DOP's functional structures help to improve the parse accuracy of tree structures. In other words, LFG-DOP outperforms Tree-DOP if evaluated on tree structures only. According to paired *t*-tests all differences in accuracy were statistically significant. This result is promising since Tree-DOP has been shown to obtain state-of-the-art performance on the Wall Street Journal corpus (see Bod 2000a).

### 5.4 Comparing Viterbi *n* best to Monte Carlo

Finally, we were interested in comparing an alternative, more efficient search method for estimating the most probable analysis. In the following set of experiments we use a Viterbi *n* best search heuristic (as explained in section 4), and let *n* range from 1 to 10,000 derivations. We also compute the results obtained by Monte Carlo for the same number of derivations. We used the same training/test set splits as in the previous experiments and used both ungeneralized and generalized fragments up to depth 4 together with the discounted RF estimator.

| Nr. of derivations | Viterbi *n* best | Monte Carlo |
|---|---|---|
| 1 | 74.8% | 20.1% |
| 10 | 75.3% | 36.7% |
| 100 | 77.5% | 67.0% |
| 1,000 | 77.5% | 77.1% |
| 10,000 | 77.5% | 77.5% |

Table 7. Precision on the Verbmobil

| Nr. of derivations | Viterbi *n* best | Monte Carlo |
|---|---|---|
| 1 | 75.6% | 25.6% |
| 10 | 76.2% | 44.3% |
| 100 | 79.1% | 74.6% |
| 1,000 | 79.8% | 79.1% |
| 10,000 | 79.8% | 80.0% |

Table 8. Precision on the Homecentre

The tables show that Viterbi *n* best already achieves a maximum accuracy at 100 derivations (at least on the Verbmobil corpus) while Monte Carlo needs a much larger number of derivations to obtain these results. On the Homecentre corpus, Monte Carlo slightly outperforms Viterbi *n* best at 10,000 derivations, but these differences are not statistically significant. Also remarkable are the relatively high results obtained with Viterbi *n* best if only one derivation is used. This score corresponds to the analysis generated by the most probable (valid) derivation. Thus Viterbi *n* best is a promising alternative to Monte Carlo resulting in a speed up of about two orders of magnitude.

## 6 Conclusion

We presented a parser which analyzes new input by probabilistically combining fragments from LFG-annotated corpora into new analyses. We have seen that the parse accuracy increased with increasing fragment size, and that LFG's functional structures contribute to significantly higher parse accuracy on tree structures. We tested two search techniques for the most probable analysis, Viterbi *n* best and Monte Carlo. While these two techniques achieved about the same accuracy, Viterbi *n* best was about 100 times faster than Monte Carlo.